\documentclass{article}
\usepackage{float}

\usepackage[preprint]{neurips_2025}

\usepackage[utf8]{inputenc} 
\usepackage[T1]{fontenc}    
\usepackage{hyperref}       
\usepackage{url}            
\usepackage{booktabs}       
\usepackage{amsfonts}       
\usepackage{nicefrac}       
\usepackage{microtype}      
\usepackage{xcolor}         
\usepackage{graphicx}
\usepackage{amsmath} 
\usepackage{pifont}
\usepackage{multirow}
\title{StoryBench: A Dynamic Benchmark for Evaluating Long-Term Memory with Multi Turns}

\author{
  Luanbo Wan${}^{1,2}$\thanks{This work was completed during Luanbo Wan's internship at Institute for AI Industry Research (AIR), Tsinghua University, Beijing, China.},\\
  {\bf Weizhi Ma${}^{1}$}\thanks{~ Corresponding author.}\\
  ${}^{1}$Institute for AI Industry Research (AIR), Tsinghua University, Beijing, China \\
  ${}^{2}$University of Electronic Science and Technology of China, Chengdu, China \\
  \texttt{mawz@tsinghua.edu.cn} \\
}

\begin{document}
\maketitle
\begin{abstract}
Long-term memory (LTM) is essential for large language models (LLMs) to achieve autonomous intelligence in complex, evolving environments. Despite increasing efforts in memory-augmented and retrieval-based architectures, there remains a lack of standardized benchmarks to systematically evaluate LLMs' long-term memory abilities. Existing benchmarks still face challenges in evaluating knowledge retention and dynamic sequential reasoning, and in their own flexibility, all of which limit their effectiveness in assessing models' LTM capabilities. To address these gaps, we propose a novel benchmark framework based on interactive fiction games, featuring dynamically branching storylines with complex reasoning structures. These structures simulate real-world scenarios by requiring LLMs to navigate hierarchical decision trees, where each choice triggers cascading dependencies across multi-turn interactions. Our benchmark emphasizes two distinct settings to test reasoning complexity: one with immediate feedback upon incorrect decisions, and the other requiring models to independently trace back and revise earlier choices after failure. As part of this benchmark, we also construct a new dataset designed to test LLMs’ LTM within narrative-driven environments. We further validate the effectiveness of our approach through detailed experiments. Experimental results demonstrate the benchmark's ability to robustly and reliably assess LTM in LLMs.
\end{abstract}

\section{Introduction}
In the field of artificial intelligence, the pursuit of true intelligence in large language models (LLMs) has prompted researchers to look to biology for inspiration~\citep{NEURIPS2024_6ddc001d,2025arXiv250415965W}. Just as organisms gradually accumulate knowledge through experience over time, LLMs need to possess long-term memory (LTM) capabilities to achieve self-evolution and strategic optimization in ever-changing environments~\citep{2025arXiv250402441S}. Moreover, as LLMs are increasingly applied in scenarios such as multi-session dialogue~\citep{zhang2025survey}, task planning, and lifelong learning, the need for models to retain, update, and leverage prior knowledge dynamically becomes critical. Without robust LTM, AI systems are limited to short-term reasoning and static knowledge use, failing to achieve sustained, autonomous intelligence.

Given the importance of LTM in enabling advanced behaviors, it is crucial to evaluate these capabilities reliably and systematically. However, current benchmarks face challenges in adequately evaluating LTM capabilities in two critical dimensions: 1) \textbf{Knowledge Retention}: the capacity to absorb, integrate, and preserve information across extended texts, maintaining contextual continuity beyond mere fact retrieval or local recall~\citep{2025arXiv250523596G, educateme_knowledge_retention}; and 2) \textbf{Sequential Reasoning}: the ability to understand and reason about sequences of events, which involves inferring latent state changes, causal dependencies, and goal shifts across complex, dynamic, and multi-turn interactions rather than simply locating pre-stated answers within static text. 3) \textbf{Flexibility}: previous benchmarks often face challenges in adjusting and evaluating in different contexts.

To address these limitations, we propose a dynamic benchmark framework inspired by interactive fiction games, where LLMs engage in branching narratives with multi-turns that simulate long-term sequential decision-making. In our benchmark, the model continuously receives scene descriptions, dialogues, and options, and must make choices based on its understanding. We design two modes: Immediate Feedback provides immediate feedback when the model makes a wrong choice, while Self Recovery allows the story to continue toward a failure ending without any hint, requiring the model to identify and revise past decisions on its own. Through this setup, our benchmark effectively evaluates the model’s ability to remember key information (knowledge retention) and reason over event sequences (sequential reasoning). Furthermore, our benchmark demonstrates excellent flexibility in accommodating diverse scenarios.

To further illustrate the advantages of our benchmark, we comprehensively evaluate the differences between existing benchmarks and ours (Table~\ref{tab:comparison}) based on the following aspects: 

\paragraph{\textbf{Knowledge Retention.}} \textbf{Long-context (L-ctx)}
evaluates whether the task requires long-term memory of earlier context to succeed. \textbf{Continuity (Conty)} measures whether the benchmark requires the model to maintain a coherent understanding of entities, events, and their relationships across interactions. 

\paragraph{\textbf{Sequential Reasoning.}} \textbf{Complexity (Comp.)} indicates whether the benchmark features non-linear reasoning tasks, where multiple interdependent events or decisions must be jointly considered, requiring the model to reason beyond sequential context. \textbf{Dynamics (Dyn.)} refers to whether the model's actions or responses influence future tasks or states in the environment. 
\textbf{Multi-turn (M-turn)} evaluates whether the task involves multiple sequential interactions, where each turn is temporally connected to the previous ones.

\paragraph{\textbf{Flexibility.}} \textbf{Multi-solution (M-sol)} indicates whether the benchmark includes tasks or questions with multiple valid answers or approaches, rather than a single fixed solution. \textbf{LTM+STM} evaluates the combined usage of long-term memory (LTM) and short-term memory (STM), i.e., whether the task requires reasoning over both recent and distant information.

\begin{table}[ht]
\caption{Comparison of Existing Benchmarks across Multiple Dimensions.}
\centering
\label{tab:comparison}
\scriptsize
\begin{tabular}{|l|c|c|c|c|c|c|c|c|}
\hline
\textbf{Benchmark} & \textbf{Type} & \multicolumn{2}{c|}{\textbf{Knowledge Retention}} & \multicolumn{3}{c|}{\textbf{Sequential Reasoning}} & \multicolumn{2}{c|}{\textbf{Flexibility}} \\
\cline{3-9}
 &  & \textbf{L-ctx} & \textbf{Conty} & \textbf{Comp.} & \textbf{Dyn.} & \textbf{M-turn} & \textbf{M-sol} & \textbf{LTM+STM} \\
\hline
NeedleInAHaystack & Synthetic & \ding{51} & \ding{55} & \ding{55} & \ding{55} & \ding{55} & \ding{55} & \ding{55} \\
RULER             & Synthetic & \ding{51} & \ding{55} & \ding{51} & \ding{55} & \ding{55} & \ding{55} & \ding{55} \\
LTM Benchmark     & Synthetic & \ding{51} & \ding{55} & \ding{51} & \ding{55} & \ding{51} & \ding{55} & \ding{51} \\
BABILong          & Synthetic & \ding{51} & \ding{55} & \ding{55} & \ding{55} & \ding{55} & \ding{55} & \ding{51} \\
L-Eval            & Realistic & \ding{51} & \ding{55} & \ding{55} & \ding{55} & \ding{55} & \ding{51} & \ding{51} \\
LongBench         & Hybrid    & \ding{51} & \ding{55} & \ding{55} & \ding{55} & \ding{55} & \ding{55} & \ding{51} \\
LooGLE            & Hybrid     & \ding{51} & \ding{55} & \ding{51} & \ding{55} & \ding{55} & \ding{55} & \ding{55} \\
InfiniteBench     & Hybrid    & \ding{51} & \ding{55} & \ding{55} & \ding{55} & \ding{55} & \ding{51} & \ding{51} \\
Ours(StoryBench)  & Hybrid  & \ding{51} & \ding{51} & \ding{51} & \ding{51} & \ding{51} & \ding{51} & \ding{51} \\
\hline
\end{tabular}
\end{table}

To validate the effectiveness of our benchmark, we conduct systematic evaluations on advanced four LLMs. Each model is tested under both evaluation modes across 80+ branching story paths, with performance measured in terms of correct decision rates, task success counts, etc. Results show that while GPT-4o~\citep{openai_gpt_4o} and Claude 3.5 Sonnet~\citep{anthropic_claude_3_5_sonnet} demonstrate relatively stronger long-term knowledge retention and sequential reasoning, all models struggle with self-recovery and fail to consistently revise earlier mistakes. In-depth failure analysis further reveals distinct memory bottlenecks, which existing benchmarks could be enhanced to expose. These findings confirm the utility of StoryBench in capturing LTM deficiencies, offering a more granular and realistic assessment than prior benchmarks.

Our contributions are as follows:
\begin{itemize}
    \item \textbf{A Dynamic Multi-turn Evaluation Framework:} We introduce a novel dynamic multi-turn benchmark inspired by interactive fiction. Through branching narratives and two distinct modes (Immediate Feedback, Self Recovery), it assesses models’ knowledge retention and sequential reasoning, while offering high flexibility across various scenarios.

    \item \textbf{A Novel Dataset for Long-Term Memory Evaluation:} We construct an annotated interactive fiction-based dataset to test LTM. It features cohesive narrative continuity, dynamic branching, complex interdependencies, and multi-solution mechanisms to emulate real-world memory challenges.
    
    \item \textbf{Reliable and Robust Experimental Analysis:} To ensure the credibility of our findings, we perform repeated trials, enhancing statistical robustness and supporting meaningful performance comparisons.
\end{itemize}

\section{Related Work}

\subsection{Strategies and Techniques for Enhancing Long-Term Memory}
Transformer models face inherent limitations in processing long sequences due to the quadratic complexity of self-attention mechanisms. To address these challenges, various architectural innovations have been proposed, including sparse attention mechanisms like Reformer ~\citep{kitaevreformer}, Longformer ~\citep{beltagy2020longformer}, Sparse Transformer ~\citep{child2019generating}, and Sparse Flash Attention ~\citep{pagliardini2023faster}, which reduce the number of token pairs in attention computation to improve speed and memory usage. 
Enhancements such as dilated convolution and cascading attention ~\citep{ding2023cervical}, sparse attention and HSR data structures ~\citep{chen2024hsr}, and Ring Attention ~\citep{liuringattention} aid in handling long-range dependencies, while performance optimizations like FlashAttention ~\citep{dao2022flashattention} and PagedAttention ~\citep{kwon2023efficient} further enhance efficiency through techniques like tiling, paging and flexible KV cache sharing. 
Context expansion techniques via recurrence, such as Transformer-XL~\citep{dai2019transformer}, enable the retention and reuse of longer context windows and optimizations in Transformer-XL, such as reducing the number of long-range memories and limiting attention range in lower layers~\citep{rae2020transformers}, can achieve comparable or better performance. 
New architectures like Mamba ~\citep{gu2024mamba} and RWKV ~\citep{peng2023rwkv} explore alternatives to traditional attention mechanisms. 
Beyond these architectural improvements, recent research has also focused on explicit memory mechanisms to address the limitations of fixed-size context windows and improve information retention. 
Memory modules like MemoryBank~\citep{zhong2023memorybank} and Retrieval-Augmented Generation (RAG) leverage external storage and dynamic retrieval to enhance long-term memory and knowledge utilization in language models.
Parameter-efficient fine-tuning techniques such as LoRA ~\citep{han2024layer} embed task-specific adjustments via low-rank matrices, preserving information without altering the full model architecture. 
Hybrid memory systems like MemGPT ~\citep{packer2023memgpt}, Mem0 ~\citep{chhikara2025mem0}, and MemoryScope~\citep{MemoryScope} integrate various memory modules and interfaces to enhance long-term retention and retrieval, while recursive approaches like Gist Memory~\citep{2024arXiv240209727L} compress and retain key context fragments. 
These advancements collectively address both retention and adaptive management of information, enabling more effective long-term memory capabilities in language models.

\subsection{Benchmarks for Evaluating Long-Term Memory}
Recent evaluations of large language models' (LLMs) long-context and long-term memory capabilities have primarily relied on dedicated benchmarks. Existing benchmarks use prefilled contexts of varying lengths, such as up to 10k tokens in LongBench~\citep{bai2024longbench}, $\approx$24k in LooGLE~\citep{li2023loogle}, and up to 100k in InfiniteBench~\citep{zhang2024bench} and even longer contexts(10 million tokens or even more) in BabiLong~\citep{kuratov2024babilong}. 
Benchmarks like ZeroSCROLLS~\citep{shaham2023zeroscrolls}, L-Eval~\citep{an2024eval}, and LongBench diversify task types and cover various domains and sequence lengths. ZeroSCROLLS focuses on zero-shot evaluation, L-Eval provides diverse long-document tasks, and LongBench spans six major task categories across multiple languages.
Synthetic retrieval-focused setups like Needle-in-a-Haystack~\citep{needle-in-haystack} are popular for their controllability, but concerns remain about their ecological validity due to overly repetitive haystacks. 
Other benchmarks like RULER~\citep{hsieh2024ruler} and BAMBOO~\citep{dong-etal-2024-bamboo} assess reasoning under long contexts. 
For a holistic understanding of long texts, ChapterBreak~\citep{sun2022chapterbreak} has been proposed. For long-range discourse modeling in multi-session conversations, Multi-Session Chats~\citep{xu2022long} has been introduced. 
Agent-based evaluations such as AgentBench~\citep{liu2024agentbench}, WebArena~\citep{zhouwebarena}, and LLF-Bench~\citep{cheng2024llf} offer dynamic environments for long-term interactions, focusing on multi-turn reasoning, real-world task completion, and learning from language feedback, respectively.
While most of these works evaluate functional behavior, few explicitly isolate long-memory capabilities. A notable exception is LTM benchmark~\citep{castillobeyond}, which targets long-term memory in multi-turn conversations. 
However, existing benchmarks still face challenges in several aspects, especially in the evaluation of knowledge retention, sequential reasoning, and flexibility.

\section{StoryBench}

\subsection{Motivation and Overview}

Existing benchmarks apply static tasks (factual recall or isolated chain of thought tasks) that do not fully capture the dynamic nature of real-world interactions~\citep{chang2024survey}, suggesting there is room for improvement in evaluating LTM abilities in two critical dimensions: \textbf{knowledge retention} and \textbf{sequential reasoning}, as well as in their own \textbf{flexibility}. The limitation of current benchmarks results from their inability to simulate the dynamic, sequential nature of real-world decision-making, where memory must be actively updated, integrated with new information, and adapted to evolving contexts through multi-turn interactions. 

To address this, we introduce StoryBench. The core design principle of StoryBench is to conduct memory stress-tests within a {dynamic} and sequentially structured environment grounded in interactive fiction {multi-turn} game-play. Unlike traditional benchmarks relying on static inputs or isolated memory recalls, StoryBench simulates realistic decision-making by embedding models in evolving narratives where each choice not only compels models to integrate information across short-term and long-term contexts ({knowledge retention}) but also tracks changing relationships between story elements and resolves contradictions arising from prior decisions in {multi-turn} interactions({sequential reasoning}). In summary, StoryBench provides a more comprehensive and dynamic framework for evaluating long-term memory capabilities, effectively enhancing the assessment of {knowledge retention} and {sequential reasoning}, as well as improving the {flexibility} of the evaluation process.

\subsection{Dynamic Narrative and Multi-Turn Decision-Making}

StoryBench leverages the inherently {dynamic} and {multi-turn} nature of interactive fiction games to assess memory in realistic decision-making trajectories. Each run through the benchmark involves a sequence of interconnected choices, where past actions shape future outcomes. The model must continuously track character states, causal dependencies, and branching outcomes over extended contexts. This setup naturally embodies several key properties:

\begin{itemize}
\item \textbf{Long-term}: Many decisions require recalling events or facts introduced a few turns earlier. Concrete examples of such dependencies are provided in Section~\ref{sec:dataset}.
\item \textbf{Continuity}: The benchmark follows a coherent plot, ensuring semantic continuity across interactions.
\item \textbf{Complex}: Consecutive decisions are not isolated, but closely linked. One choice may directly affect the conditions or outcomes of several subsequent ones. We provide detailed illustrations of such dependencies in Section~\ref{sec:dataset}.
\item \textbf{Dynamic}: Incorrect or suboptimal decisions dynamically alter the story path or trigger failure endings, requiring the model to adapt in real-time.
\item \textbf{Multi-turn}: The task unfolds over many turns, demanding sustained memory and reasoning across sequentially extended interactions.
\item \textbf{Multi-solution}: Many decision points allow for multiple acceptable paths, rather than a single fixed correct answer, better reflecting the uncertainty and variability of real-world scenarios. Specific examples demonstrating the multi-solution nature of the benchmark are provided in Section~\ref{sec:dataset}.
\end{itemize}

\subsection{Two Task Modes for Evaluating LTM}

To explore different aspects of memory utilization, we design two complementary task modes. The dual-mode setup allows StoryBench to probe both short-horizon reactive memory and long-horizon strategic recall (LTM), offering a comprehensive view of how models navigate extended, decision-heavy interactions and revealing not just whether a model can remember facts, but whether it can strategically reason across time, self-correct, and navigate branching storylines over extended sequences.

\begin{figure}[ht] 
    \centering 
    \includegraphics[width=0.87\textwidth]{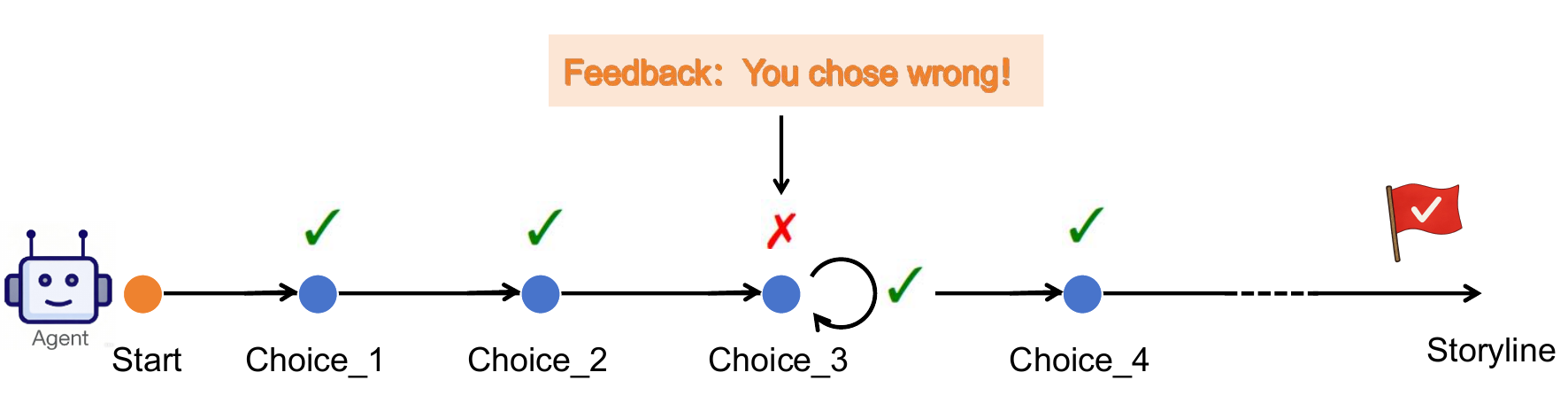} 
    \caption{Immediate Feedback. The model is informed immediately after each incorrect choice and prompted to retry until the correct option is selected.} 
    \label{fig:immediate} 
\end{figure}

\textbf{Immediate Feedback:} Designed to evaluate a model's responsiveness to error signals, this mode simulates situations where feedback is available at each turn. After a wrong choice, the model is told the outcome and prompted to retry (Figure~\ref{fig:immediate}), allowing us to examine its short-term adjustment ability and interactive learning dynamics.

\begin{figure}[ht] 
    \centering 
    \includegraphics[width=0.9\textwidth]{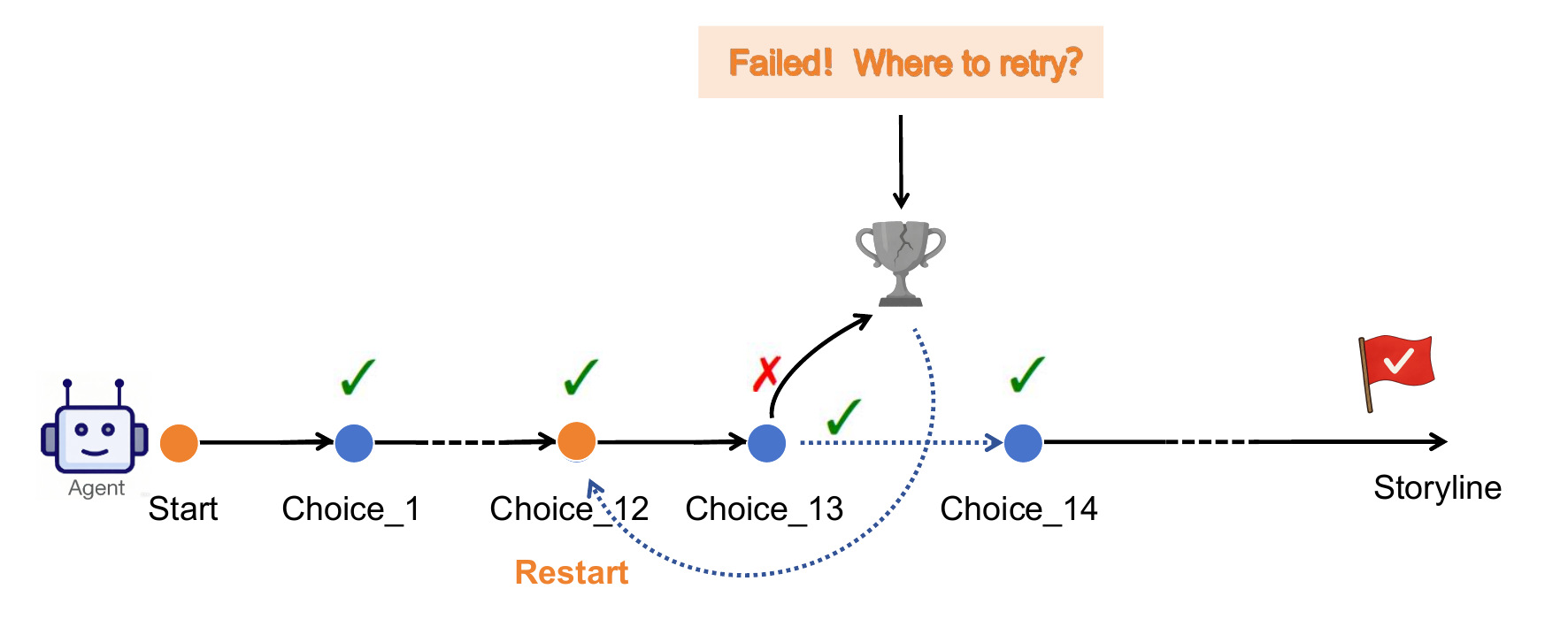} 
    \caption{Self Recovery. An incorrect choice leads to a failure ending either immediately or after several scenes. The model is then asked to identify the earliest point in the story where it believes the incorrect decision occurred and to attempt recovery from that point.} 
    \label{fig:self} 
\end{figure}

\textbf{Self Recovery:} This mode suppresses feedback, mimicking scenarios where incorrect decisions propagate through multiple scenes, potentially ending the game. The model is then challenged to trace back to the error’s origin and recover (Figure~\ref{fig:self}). This stresses long-term causal reasoning and memory retention under uncertainty.

\subsection{Tailored Metrics for Assessing LTM Models}

To comprehensively evaluate long-term memory (LTM) capabilities in language models, StoryBench introduces a set of targeted metrics covering two essential cognitive dimensions: knowledge retention and sequential reasoning.

We define a decision sequence $\{c_1, c_2, ..., c_T\}$, where $c_t \in \{0, 1\}$ denotes whether the model selected the correct option ($1$) or not ($0$) at step $t$.

\subsubsection{Metrics for Knowledge Retention}

\begin{itemize}
    \item \textbf{Overall Accuracy (Overall Acc)}:  
    The average correctness across all decisions, measuring how consistently the model maintains relevant knowledge and narrative coherence:
    \[
    \text{Accuracy}_{\text{overall}} = \frac{1}{T} \sum_{t=1}^{T} c_t.
    \]

    \item \textbf{First-Try Accuracy (First-Try Acc)}:  
    The proportion of decision points at which the model selected the correct option on its first attempt. Let $f_t \in \{0,1\}$ be $1$ if the model is correct on the first try at step $t$, then:
    \[
    \text{Accuracy}_{\text{first-try}} = \frac{1}{T} \sum_{t=1}^{T} f_t.
    \]

    \item \textbf{Longest Consecutive Correct Sequence (Longest Corr)}:  
    The length of the longest contiguous subsequence of correct decisions:
    \[
    \text{LongestCorr} = \max_{1 \leq i \leq j \leq T} \left( j - i + 1 \ \big| \ c_k = 1 \ \forall k \in [i, j] \right).
    \]
    This reflects the model’s ability to sustain contextual consistency over extended intervals, though less critical than the above metrics.
\end{itemize}

\subsubsection{Metrics for Sequential Reasoning}

\begin{itemize}
    \item \textbf{Accuracy by Difficulty (Easy/Hard Acc)}: To account for varying levels of memory and reasoning demand, we classify decisions into \textit{easy} and \textit{hard} categories. A decision is labeled as \textit{hard} if it requires recalling information from a distant context, tracking latent state changes, or performing multi-step sequential reasoning; otherwise, it is considered \textit{easy}. Let $\mathcal{E}_t$ and $\mathcal{H}_t$ denote easy and hard decision sets up to step $t$ (including retries), then:
    \[
    \text{Accuracy}_{\text{easy}}^{(t)} = \frac{1}{|\mathcal{E}_t|} \sum_{i \in \mathcal{E}_t} c_i, \quad \text{Accuracy}_{\text{hard}}^{(t)} = \frac{1}{|\mathcal{H}_t|} \sum_{i \in \mathcal{H}_t} c_i.
    \]
    These metrics assess how well the model adapts to sequentially distributed and cognitively demanding decisions.

    \item \textbf{Retry Count}:  
    Let $r_t$ denote the number of retries required before reaching a correct decision at step $t$. The total number of retries across the trajectory is:
    \[
    \text{Retry}_{\text{total}} = \sum_{t=1}^{T} r_t.
    \]

    \item \textbf{Max Error per Choice (Max Err/Choice)} and \textbf{Thresholded Error Count}:  
    These metrics capture the worst-case and accumulated difficulty for the model in terms of repeated failures:
    \[
    \text{MaxError} = \max_{1 \leq t \leq T} r_t, \quad 
    \text{ErrorCount}_{\geq r_{\text{thres}}} = \sum_{t=1}^{T} \mathbb{I}(r_t \geq r_{\text{thres}}),
    \]
    Where $\mathbb{I}(\cdot)$ is the indicator function and $r_{\text{thres}}$ is a predefined retry threshold (e.g., 9 in our experiments).
\end{itemize}

Finally, while not directly measuring memory accuracy, two auxiliary metrics provide additional perspective on the model’s efficiency in handling long-horizon tasks: \textbf{Runtime Cost} reflects the inference efficiency of the memory system, while \textbf{Token Consumption (Token Cons)} indicates the model’s reliance on contextual information.

Together, these metrics form a multi-faceted evaluation framework that jointly targets both the persistence of stored information and the model’s ability to apply it dynamically within complex, sequentially structured environments. This ensures that memory is not only retained but also meaningfully used to navigate and reason through realistic multi-turn interactions.

\section{Dataset Construction}

\subsection{Overview}

To evaluate long-term memory (LTM) capabilities of large language models (LLMs), we construct a narrative dataset based on the interactive fiction game \textit{The Invisible Guardian}, encompassing 311 scene nodes and 86 choice nodes as captured in our structured JSON format. 

We chose to use an interactive fiction game as the basis for our dataset rather than synthetic data or real-world data for several reasons. First, it is arguable that all publicly available benchmark test cases might occasionally be included in LLM pre-training data~\citep{2024arXiv240407503L}. Consequently, to mitigate potential data overlap issues, we opted to independently construct a dataset of interactive fiction games. Second, synthetic data is often overly simplistic and lacks the nuanced coherence of real human narratives~\citep{2024arXiv240101629H}. It relies on predefined templates, resulting in repetitive scenarios that fail to capture the complex interdependencies crucial for evaluating long-term reasoning. In contrast, the interactive fiction game \textit{The Invisible Guardian} offers a rich, evolving storyline that naturally tests long-term dependencies. Third, real-world data is messy and difficult to control~\citep{2025arXiv250107487X, Behr2025}. It is influenced by numerous external factors, making it hard to isolate causal relationships and define clear ``success'' or ``failure'' paths. The structured and controlled environment of an interactive fiction game provides a clear framework for evaluating long-term memory and decision-making in a repeatable manner.

Our design incorporates several distinctive features for evaluating LTM. First, unlike conventional QA or dialogue datasets that consist of isolated or short-context samples, our dataset presents a continuous and evolving story world that unfolds over multiple interactive turns, offering a naturalistic setting for evaluating long-horizon reasoning. Second, many long-term choices depend on events or facts introduced several turns earlier, thereby testing models' long-term dependency tracking. Third, the story dynamically evolves based on the model's choices, allowing branching into different paths, including success or failure endings. Fourth, the benchmark reflects realistic decision-making complexity: consecutive choices are often interdependent, requiring models to maintain logical consistency across transitions. Finally, the dataset is multi-solution: multiple choice paths may lead to successful conclusions, emphasizing adaptability rather than rigid answer matching. 

\subsection{Structural Organization}
\label{sec:dataset}
The dataset is organized as a directed acyclic graph (DAG) composed of two types of nodes: \textit{scene nodes}, which represent narrative fragments, and \textit{choice nodes}, which define branching decision points. Edges denote transitions between these nodes, forming a tree-like structure that allows non-linear progression through the story. This organization not only captures the dynamic and interactive nature while enabling clear tracing of causal dependencies but also allows flexible nuanced evaluation of LTM in knowledge retention and sequential reasoning.

\begin{figure}[ht] 
    \centering 
    \includegraphics[width=0.85\textwidth]{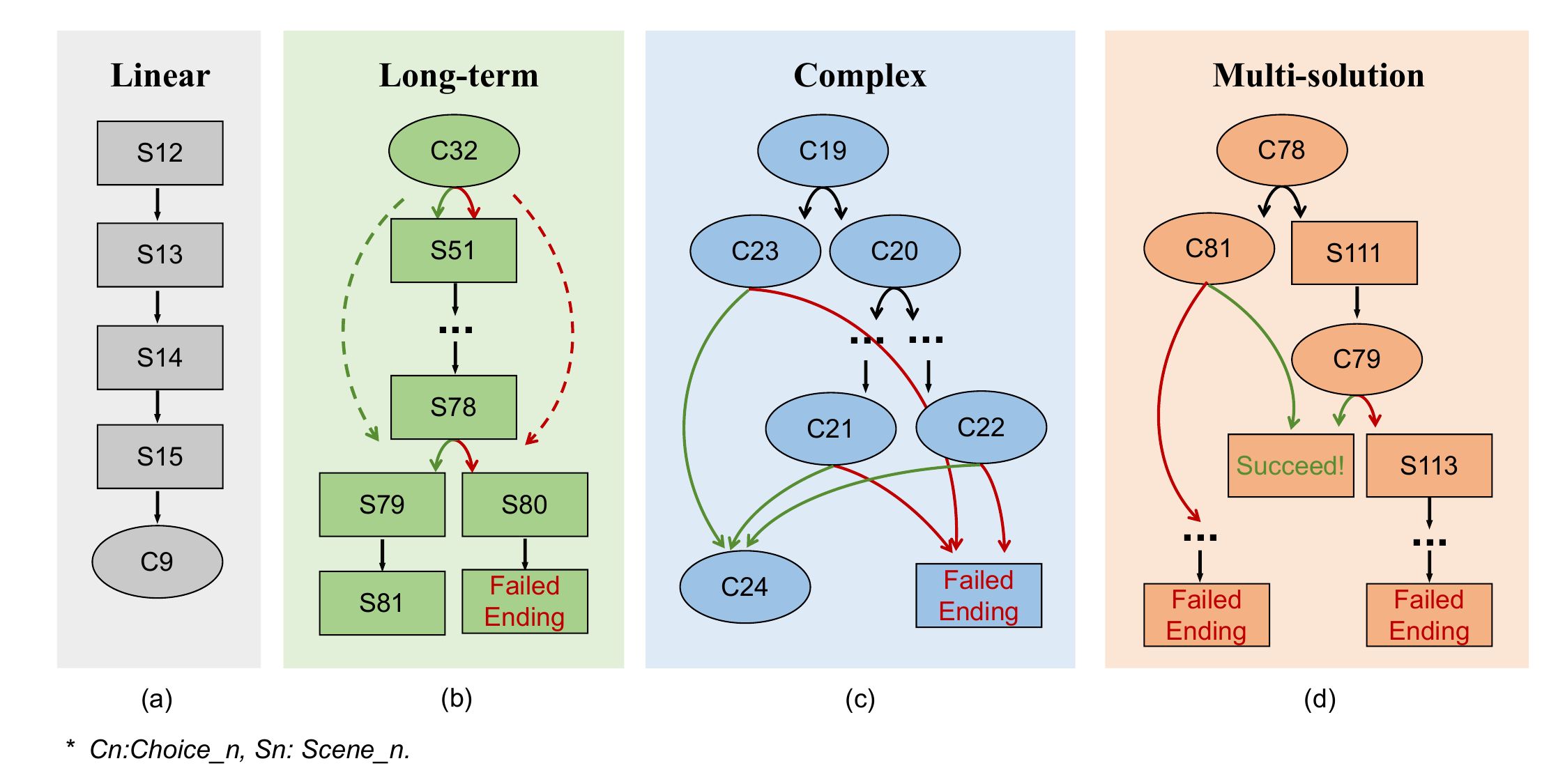} 
    \caption{Four typical patterns illustrating dataset structure complexity.} 
    \label{fig:structure} 
\end{figure}

To illustrate the complexity and diversity of our dataset's structure, we categorize representative graph patterns in Figure~\ref{fig:structure}. These include (a) \textbf{linear} chains of scenes, testing narrative understanding and short-range memory; (b) \textbf{long-term
} dependencies, where early events influence distant outcomes; (c) clusters of interdependent decisions, reflecting \textbf{complex} causal reasoning; and (d) \textbf{multi-solution} branches, where multiple paths can reach valid endings. 

\subsection{Data Source and Annotation Process}
\begin{figure}[ht] 
    \centering 
    \includegraphics[width=\textwidth]{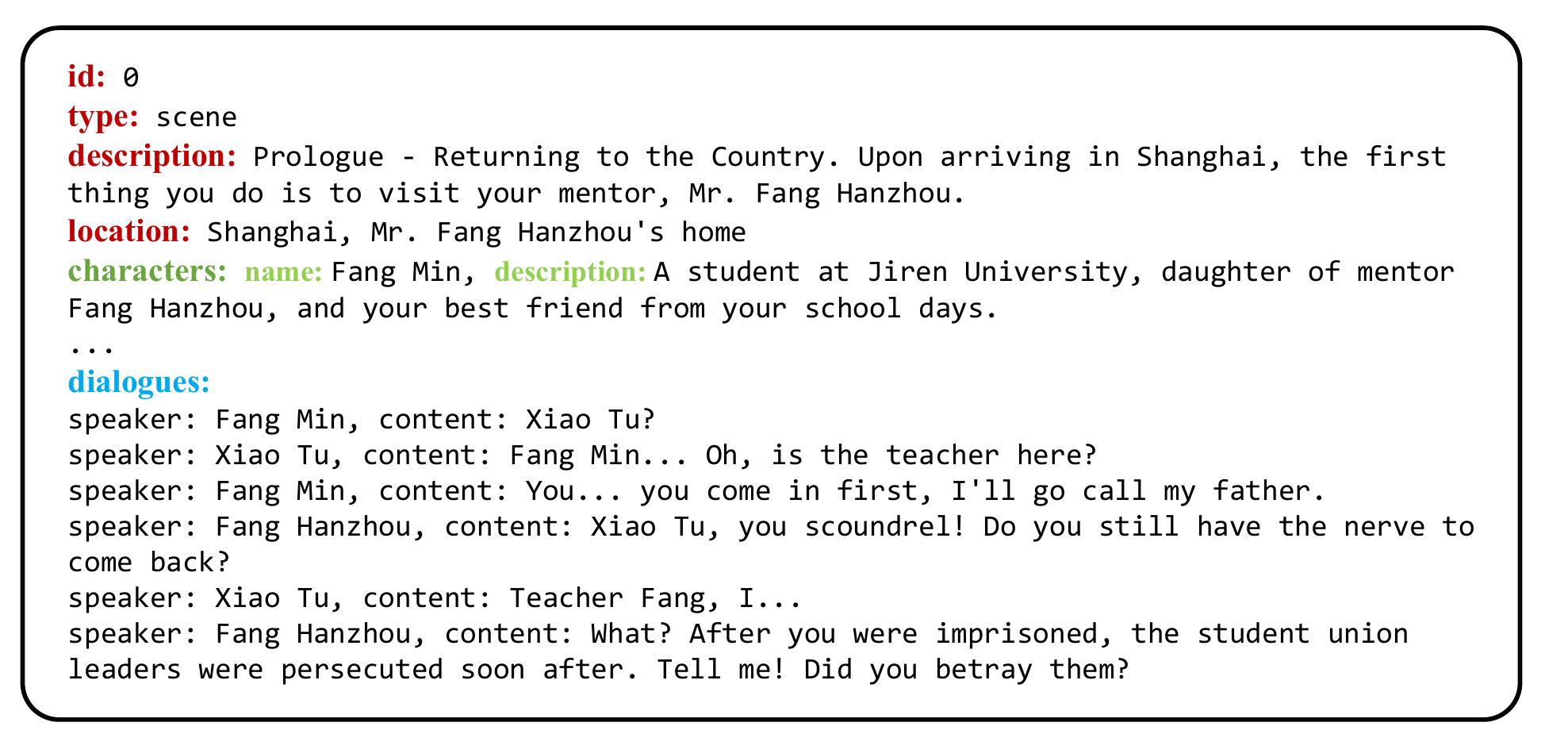} 
    \caption{Scene node example with character descriptions, dialogues, and other details.} 
    \label{fig:scene} 
\end{figure}

\begin{figure}[ht] 
    \centering 
    \includegraphics[width=0.8\textwidth]{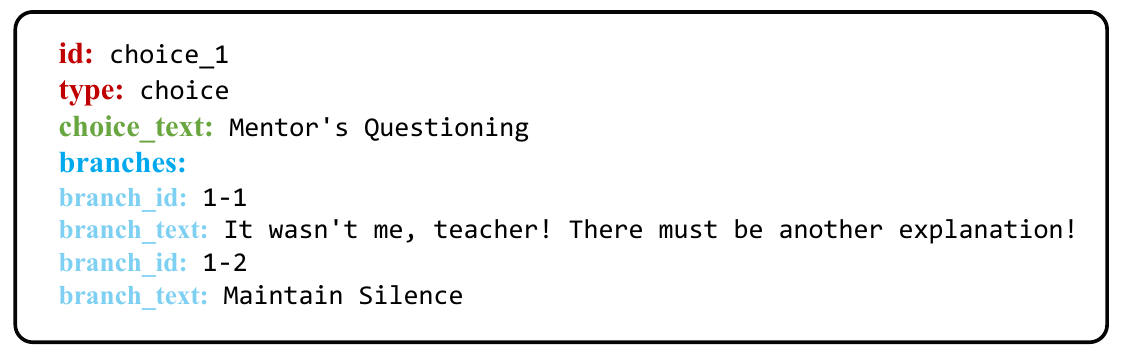} 
    \caption{Choice node example with choice text, branches, and other details.} 
    \label{fig:choice} 
\end{figure}

We construct our dataset based on the interactive fiction game The Invisible Guardian from the game's prologue to Chapter 5 by far. Manual annotation preserves the game’s branching logic and causal relationships, ensures chronological ordering with memory checkpoints, and annotates metadata on transitions, dynamics, and ethics to retain sequential depth for evaluating LLMs' long-term reasoning. All content is meticulously transcribed from the original game, encompassing dialogues, narrative descriptions, character interactions, and player decision points, with each entry structured as a JSON object annotated with granular details according to its type. Scene nodes (311 entries) include unique identifiers, location, characters with descriptive attributes, sequential dialogues with speaker labels, and flags for narrative endings (where applicable), such as ending (Figure~\ref{fig:scene}). Choice nodes (86 entries) feature unique identifiers, decision context descriptions, and branching options with distinct IDs and text (Figure~\ref{fig:choice}). 

\section{Experiments \& Results}

\subsection{Experimental Setup}

We conduct experiments on four representative foundation models: Doubao 1.5-pro-256k~\citep{doubao_1_5_pro}, GPT-4o~\citep{openai_gpt_4o}, Claude 3.5 Sonnet~\citep{anthropic_claude_3_5_sonnet}, and Deepseek-R1~\citep{2025arXiv250112948D}. These models are chosen based on both their broad real-world usage and competitive performance. 
Doubao 1.5-pro-256k excels in handling extremely long contexts with its 256k-token support, making it ideal for tasks requiring extensive context retention. 
GPT-4o, as a leading closed-source commercial model, demonstrates strong language understanding and reasoning abilities.
Claude 3.5 Sonnet excels in long-context understanding and knowledge reasoning (supporting 200k+ tokens), maintaining stable performance in long-text reasoning and structural analysis tasks.
Deepseek-R1 employs pure reinforcement learning, which gives it excellent logical reasoning and structured thinking capabilities. It shows strong performance in multi-step reasoning and planning tasks.
Their diverse features make them ideal for evaluating long-term memory across different technical approaches and application scenarios.
While several memory-augmented approaches ~\citep{chhikara2025mem0, MemoryScope} have RAG-style architectures or external memory buffers, we exclude them from our evaluation because their memory utility centers on retrieving isolated factual content. However, StoryBench emphasizes \textbf{long-term sequential reasoning}, where memory must support inference, self-correction, and causal tracking.

For each of the two task modes in StoryBench, we run 10 trials per model. Inputs are carefully formatted to encourage structured reasoning, and we adopt a Chain-of-Thought (CoT) prompting strategy to stimulate stepwise deliberation. In \textbf{Immediate Feedback} mode (results in Table~\ref{tab:if}), we observe that GPT-4o is more sensitive to content filtering issues (e.g., mentioning weapon-related terms) and frequently interrupts completion due to server overload. To ensure smooth evaluation, we filter potentially problematic vocabulary and limit single-turn inputs to 5,000 tokens for GPT-4o. In \textbf{Self Recovery }mode, models often repeatedly select the same wrong option more than ten times without real-time feedback, therefore stalling the task. So we implement a soft intervention by revealing the correct answer if a model failed at the same decision point for nine consecutive attempts. In the initial evaluation phase (results in Table~\ref{tab:sr-orig}), we retain the original unfiltered dataset and deliberately remove token limits to simulate high-pressure, long-horizon conditions, then conduct five trials. The performance of all models decreases significantly, reflecting the intrinsic difficulty of the task. In response, we launch a second phase of five-trial experiments (Table~\ref{tab:sr-improved}) with improved input handling: sensitive vocabulary is filtered and a 5,000-token per turn limit is applied. 

\subsection{Main Results of Long-Term Memory Performance}

\begin{table}[ht]
\centering
\caption{Performance of different models (Immediate Feedback).}
\label{tab:if}
\begin{tabular}{lcccc}
\toprule
\textbf{Metrics }& \textbf{Doubao1.5-pro}& \textbf{GPT-4o} & \textbf{Claude 3.5 Sonnet} & \textbf{Deepseek-R1} \\
\midrule
Overall Acc  (\%)& \textbf{80.98} ± 1.31 & 71.88 ± 1.03 & 74.86 ± 1.05 & 70.45 ± 4.62 \\
First-Try Acc  (\%)& \textbf{79.14} ± 1.33 & 63.49 ± 2.59 & 68.21 ± 1.55 & 65.16 ± 2.41 \\
Hard Acc  (\%)& \textbf{74.47} ± 2.26 & 66.94 ± 1.38 & 69.38 ± 1.26 & 60.21 ± 4.61 \\
Easy Acc  (\%)& \textbf{88.68} ± 0.15 & 77.43 ± 0.88 & 81.35 ± 1.67 & 84.94 ± 4.44 \\
Retry Count & \textbf{14.67} ± 1.25 & 24.67 ± 1.25 & 20.88 ± 1.17 & 26.40 ± 5.95 \\
Longest Corr & 10.00 ± 0.00 & 8.00 ± 0.82 & 8.50 ± 2.12 & \textbf{10.20} ± 1.47 \\
\midrule
Runtime Cost (s)& \textbf{0.65k} ± 0.02k & \underline{0.44k ± 0.08k} & 2.14k ± 0.16k & 2.72k ± 0.23k \\
Token Cons & \textbf{2043k} ± 53k & \underline{342k ± 5.8k} & 3405k ± 150k & 2396k ± 264k \\
\midrule
Success Count & 3.00 & 3.00 &\textbf{8.00} & 5.00 \\
\bottomrule
\end{tabular}
\end{table}

\begin{table}[ht]
\centering
\caption{Performance of different models (Original Self Recovery).}
\label{tab:sr-orig}
\begin{tabular}{lccccc}
\toprule
\textbf{Metrics} & \textbf{Doubao1.5-pro} & \textbf{GPT-4o} & \textbf{Claude 3.5 Sonnet} & \textbf{Deepseek-R1} \\
\midrule
Overall Acc (\%)& \textbf{69.66} & - & 68.40 ± 2.88 & - \\
First-Try Acc (\%)& \textbf{83.05} & - & 68.28 ± 1.07 & - \\
Hard Acc (\%)& 58.33 & - & \textbf{60.35} ± 4.09 & - \\
Easy Acc (\%)& \textbf{93.10} & - & 77.23 ± 0.31 & - \\
Retry Count & \textbf{21.00} & - & 21.50 ± 2.50 & - \\
Longest Corr & \textbf{15.00} & - & 13.50 ± 1.50 & - \\
Max Err/Choice & 9.00 & - & \textbf{6.00 }± 2.00 & - \\
$ \text{ErrorCount}_{\geq 9} $ & 2.00 & - & \textbf{0.00} ± 0.00 & - \\
\midrule
Runtime Cost (s)& \textbf{1.00k} & - & 3.24k ± 0.36k & - \\
Token Cons & \textbf{4158k} & - & 5532k ± 37k & - \\
\midrule
Success Count & 1.00 & 0.00 & \textbf{2.00} & 0.00 \\
\bottomrule
\end{tabular}
\end{table}

\begin{table}[ht]
\centering
\caption{Performance of different models (Improved Self Recovery).}
\label{tab:sr-improved}
\begin{tabular}{lcccc}
\toprule
\textbf{Metrics }& \textbf{Doubao1.5-pro}& \textbf{GPT-4o} & \textbf{Claude 3.5 Sonnet} & \textbf{Deepseek-R1} \\
\midrule
Overall Acc (\%)& \textbf{73.68} & 60.76 ± 1.35 &  -  & 70.18 \\
First-Try Acc  (\%)& \textbf{83.33} & 58.57 ± 1.43 &  -  & 75.41 \\
Hard Acc (\%)& 62.22 & 52.84 ± 2.06 &  -  & \textbf{62.50} \\
Easy Acc (\%)& \textbf{90.32} & 72.72 ± 2.27 &  -  & 88.24 \\
Retry Count & \textbf{17.00} & 30.50 ± 4.50 &  -  & 26.00 \\
Longest Corr & \textbf{16.00} & 8.00 ± 1.00 &  -  & 12.00 \\
Max Err/Choice  & 9.00 & \textbf{7.00} ± 2.00 &  -  & 9.00 \\
$ \text{ErrorCount}_{\geq 9} $ & 1.00 & \textbf{0.00} ± 0.00 &  -  & 2.00 \\
\midrule
Runtime Cost  (s)& 0.60k & \textbf{0.58k} ± 0.05k &  -  & 4.64k \\
Token Cons & \textbf{343k} & 510k ± 40k &  -  & 549k \\
\midrule
Success Count& 1.00 & \textbf{2.00}  & 0.00 & 1.00  \\
\bottomrule
\end{tabular}
\end{table}

\subsubsection{Model Analysis}

\begin{figure}[ht] 
    \centering 
    \includegraphics[width=\textwidth]{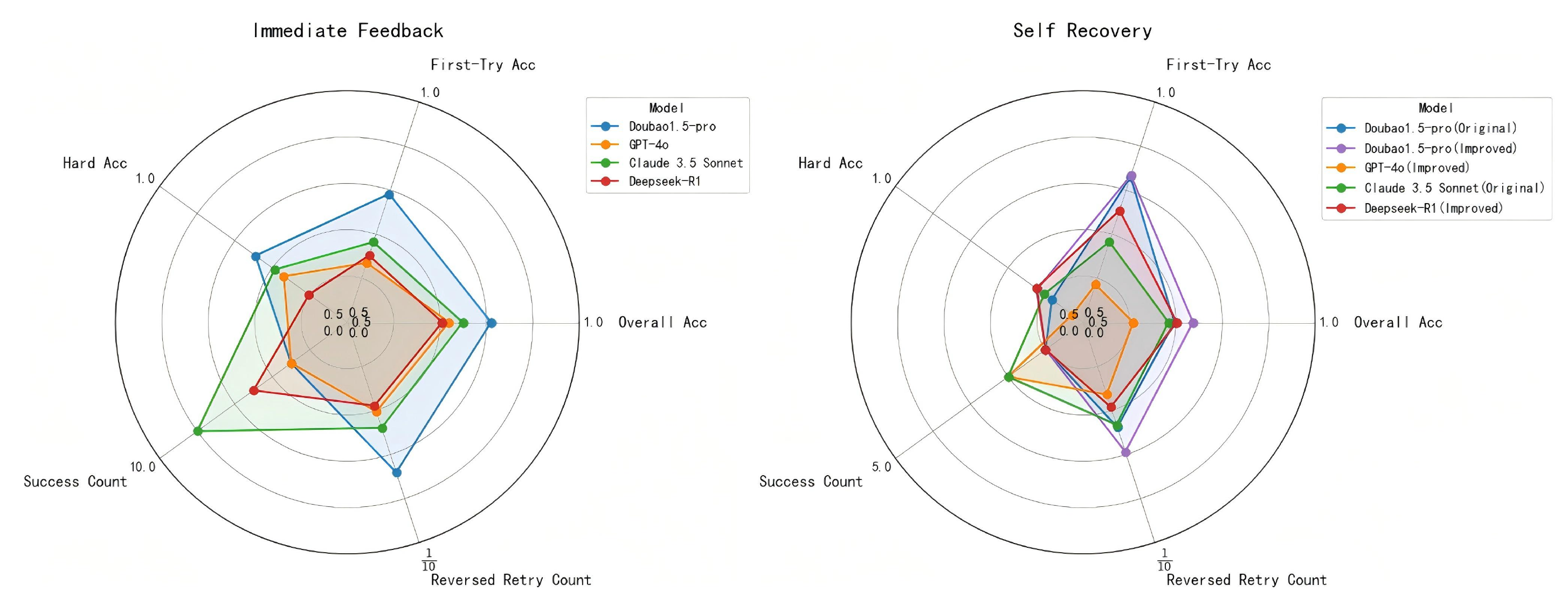} 
    \caption{Model multidimensional performance in Immediate Feedback and Self Recovery modes.} 
    \label{fig:radar} 
\end{figure}

To better understand the performance differences among models, we analyze five core metrics illustrated in Figure~\ref{fig:radar}.
\textit{Overall Accuracy} and \textit{First-Try Accuracy} reflect \textbf{knowledge retention}, capturing the model’s ability to maintain consistent and contextually grounded responses across extended interactions. \textit{Hard Accuracy} and \textit{Retry Count} assess \textbf{sequential reasoning}, as they target the model’s capacity to navigate complex, dynamic, and multi-step decision paths involving long-range dependencies. \textit{Success Count} captures the overall task-completion ability. Among all models, Doubao1.5-pro achieves the highest scores in knowledge-related metrics such as \textit{Overall Accuracy} and \textit{First-Try Accuracy}, suggesting strong capabilities in \textbf{knowledge retention}. Doubao effectively absorbs and integrates contextual information across extended texts. However, long-term memory evaluation must prioritize not accuracy but the ability to complete extended decision paths. That is because the premise for evaluation is the model can complete the story chain, otherwise, no matter how high the local accuracy is, it will lose significance to the evaluation. Doubao's \textit{Success Count} is significantly lower than Claude 3.5 Sonnet, indicating that it often "dies in details" when dealing with complex reasoning chains and long-term interactive tasks despite its solid knowledge base. In contrast, Claude 3.5 Sonnet maintains a solid balance: it trails slightly in accuracy, but excels in degree of completion, achieving the highest \textit{Success Count}. This suggests Claude is more robust in multi-turn \textbf{sequential reasoning}, which is a critical factor in long-term memory evaluation. 

\begin{figure}[ht] 
    \centering 
    \includegraphics[width=\textwidth]{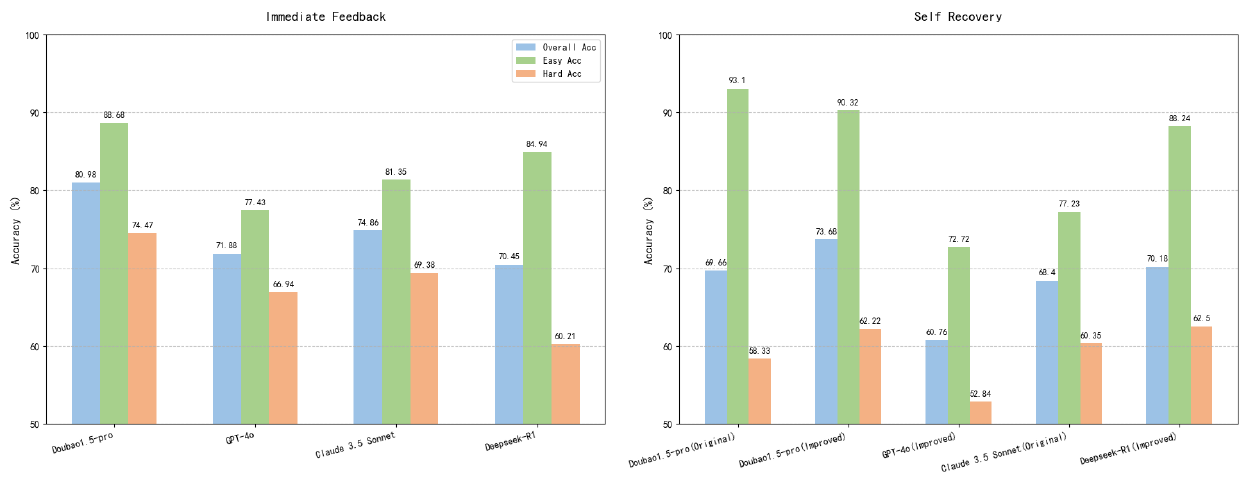} 
    \caption{Accuracy disparities across Models: overall, easy \& hard tasks.} 
    \label{fig:acc} 
\end{figure}

Interestingly, most models show large gaps between \textit{Easy} and \textit{Hard Accuracy}, Figure~\ref{fig:acc} reflecting their large gaps in \textbf{sequential reasoning}. Notably, Claude and GPT-4o show more consistent performance across difficulty levels, while Deepseek-R1, though competent in \textit{Easy Accuracy}, suffer significant drops in harder decisions, highlighting challenges in difficult or deceptive decision points that require multi-step reasoning, delayed consequences, or implicit state tracking.

From an efficiency perspective, GPT-4o and Doubao1.5-pro offer excellent cost-performance tradeoffs. Their \textit{Runtime Cost} and \textit{Token Consumption} are significantly lower than Claude 3.5 and Deepseek-R1.

\subsubsection{Insights of Distinctions Between Two Modes}

To investigate how short-term and long-term memory settings affect model behavior, we compare performance under two task modes. \textbf{Immediate Feedback} mode provides corrective signals after each wrong choice, effectively mimicking short-term memory and aiding models in adjusting quickly. In contrast, \textbf{Self Recovery} better simulates real long-term memory scenarios by removing such signals, requiring the model to navigate the narrative without external guidance.

Unsurprisingly, all models perform worse under Self Recovery mode, as shown by the consistent drop in \textit{Overall Accuracy} and \textit{Success Count}. This highlights the increased difficulty of sustained sequential reasoning and knowledge retention without short-term feedback. To alleviate task failure in extreme cases, we introduce an auxiliary intervention metric: \textit{Number of Choices Reaching Error Threshold} (we set the threshold to 9). If a model makes the same mistake more than 9 times, it is prompted with the correct answer. Only Claude 3.5 and GPT-4o never reach this threshold, suggesting that their task completions in Self Recovery mode are entirely due to self-correction and internal reasoning without any artificial hints. This contrasts sharply with other models, indicating that they excel in sustained sequential reasoning and knowledge retention.

\begin{figure}[ht] 
    \centering 
    \includegraphics[width=\textwidth]{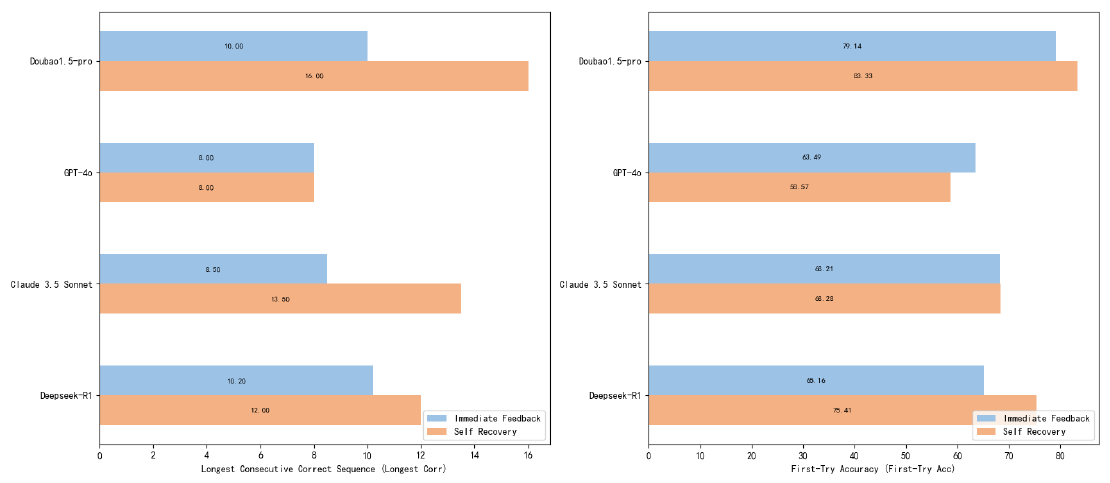} 
    \caption{Mode impact on models: First-Try Accuracy \& Longest Consecutive Correct Sequence metrics.} 
    \label{fig:surprise} 
\end{figure}

Surprisingly, despite the overall decline in performance across models in Self Recovery, two metrics: \textit{Longest Consecutive Correct Sequence} and \textit{First-Try Accuracy} actually increase for several models (Figure~\ref{fig:surprise}).
This amazing trend emphasizes that while short-term feedback aids local correction, it may also disrupt long-horizon coherence. By removing it, models foster a deeper narrative understanding (\textbf{knowledge retention}) and more coherent reasoning (\textbf{sequential reasoning}) and we better expose the true limitations and strengths of long-term memory in different models. 

A notable case is Deepseek-R1. While it does not lead in most individual metrics, it demonstrates remarkable consistency across both Immediate Feedback and Self Recovery modes. This stable performance suggests that the model is capable of making accurate revisions during backtracking.

\subsection{Failure Case Study}

In evaluating long-term memory capabilities with StoryBench, we identified two principal types of failure that reflect limitations in current language models, corresponding to the core dimensions of \textbf{knowledge retention} and \textbf{sequential reasoning}.

The most prominent issue in \textbf{knowledge retention} was the failure to preserve contextual consistency over extended narratives. Models frequently made decisions that contradicted earlier story events, character motivations, or established world logic. This suggests difficulty in integrating and maintaining distributed information over long spans of interaction, especially when the necessary context spans dozens of turns. Even when the relevant facts appeared in the prompt, models struggled to apply them coherently, indicating limitations beyond simple factual recall.

In terms of \textbf{sequential reasoning}, a critical failure case was the inability to repair long-term or multi-error decisions. In Self Recovery mode, successful completion often required models to trace errors back across multi-step causal chains and revise earlier decisions (even multiple choices in combination) that affected downstream outcomes. However, most models exhibited shallow search strategies, typically backtracking only one or two steps rather than engaging in deeper reasoning about the narrative structure or goal shifts. This myopic behavior led to persistent failure when task success depended on understanding and correcting long-term dependencies. We retained such failures to reflect the true upper-bound difficulty of long-term memory reasoning.

Other failures such as format mismatches (e.g., returning option indices instead of decision point IDs), content filtering blocks, server timeouts, or rare instances of hallucinated explanations were also observed but were comparatively infrequent. These were retained in evaluation for completeness but are not the focus of our analysis.

These diverse failure cases underscore the challenge of StoryBench and emphasize the need for more robust memory integration, format alignment, and long-range error correction in current foundation models.

\section{Limitations}
While our benchmark provides a comprehensive evaluation of long-term memory capabilities in large language models through complex, branching narrative tasks, it has several limitations. First, the scenarios are derived from a single interactive fiction domain and the interactive environment is text-based, both of which may limit the benchmark's generalizability to other knowledge-intensive or task-oriented contexts that require multimodal support. Second, the number of turns and the length of the context are still limited. The current interactive fiction dataset consists of only 6 chapters, which may not fully capture the long-term dependencies and complex reasoning required in more extensive narratives. Future work could expand the dataset by adding subsequent chapters to provide a more comprehensive evaluation of long-term memory. Third, due to API constraints and cost, we primarily evaluate a limited number of mainstream models. The performance of other models under similar conditions remains unexplored. Fourth, although we include a self-recovery setting to simulate real-world error correction, the evaluation remains scripted and cannot capture all forms of natural feedback.

\section{Conclusions}
We introduce StoryBench, a novel benchmark designed to systematically evaluate long-term memory capabilities in complex, dynamic, and multi-turn narrative environments. 
By simulating realistic memory demands across story understanding, sequential inference, and flexible correction, our benchmark assesses current mainstream models in knowledge retention and sequential reasoning.
Through comprehensive experiments on representative models and detailed failure case analyses, we demonstrate that current models exhibit significant performance gaps on our benchmark, highlighting StoryBench's difficulty and effectiveness in evaluating long-term memory capabilities. 
Our findings underscore the importance of developing more robust memory mechanisms, laying the groundwork for future research toward memory-augmented, context-aware language agents.

\small  
\bibliography{main}

\begin{thebibliography}{49}
\providecommand{\natexlab}[1]{#1}
\providecommand{\url}[1]{\texttt{#1}}
\expandafter\ifx\csname urlstyle\endcsname\relax
  \providecommand{\doi}[1]{doi: #1}\else
  \providecommand{\doi}{doi: \begingroup \urlstyle{rm}\Url}\fi

\bibitem[An et~al.(2024)An, Gong, Zhong, Zhao, Li, Zhang, Kong, and Qiu]{an2024eval}
Chenxin An, Shansan Gong, Ming Zhong, Xingjian Zhao, Mukai Li, Jun Zhang, Lingpeng Kong, and Xipeng Qiu.
\newblock L-eval: Instituting standardized evaluation for long context language models.
\newblock In \emph{Proceedings of the 62nd Annual Meeting of the Association for Computational Linguistics (Volume 1: Long Papers)}, pages 14388--14411, 2024.

\bibitem[Anthropic(2024)]{anthropic_claude_3_5_sonnet}
Anthropic.
\newblock Claude 3.5 sonnet model card addendum, 2024.
\newblock URL \url{https://www.anthropic.com/news/claude-3-5-sonnet}.

\bibitem[Bai et~al.(2024)Bai, Lv, Zhang, Lyu, Tang, Huang, Du, Liu, Zeng, Hou, et~al.]{bai2024longbench}
Yushi Bai, Xin Lv, Jiajie Zhang, Hongchang Lyu, Jiankai Tang, Zhidian Huang, Zhengxiao Du, Xiao Liu, Aohan Zeng, Lei Hou, et~al.
\newblock Longbench: A bilingual, multitask benchmark for long context understanding.
\newblock In \emph{Proceedings of the 62nd Annual Meeting of the Association for Computational Linguistics (Volume 1: Long Papers)}, pages 3119--3137, 2024.

\bibitem[Behr et~al.(2025)Behr, Burghaus, Diedrich, et~al.]{Behr2025}
Matthias Behr, Ralf Burghaus, Christoph Diedrich, et~al.
\newblock Opportunities and challenges for ai-based analysis of rwd in pharmaceutical r\&d: A practical perspective.
\newblock \emph{Künstliche Intelligenz}, 39\penalty0 (1):\penalty0 7--18, 2025.
\newblock \doi{10.1007/s13218-023-00809-6}.
\newblock URL \url{https://doi.org/10.1007/s13218-023-00809-6}.

\bibitem[Beltagy et~al.(2020)Beltagy, Peters, and Cohan]{beltagy2020longformer}
Iz~Beltagy, Matthew~E Peters, and Arman Cohan.
\newblock Longformer: The long-document transformer.
\newblock \emph{arXiv preprint arXiv:2004.05150}, 2020.

\bibitem[ByteDance(2025)]{doubao_1_5_pro}
ByteDance.
\newblock Doubao-1.5-pro, 2025.
\newblock URL \url{https://seed.bytedance.com/zh/special/doubao_1_5_pro}.

\bibitem[Castillo-Bolado et~al.(2024)Castillo-Bolado, Davidson, Gray, and Rosa]{castillobeyond}
David Castillo-Bolado, Joseph Davidson, Finlay Gray, and Marek Rosa.
\newblock Beyond prompts: Dynamic conversational benchmarking of large language models.
\newblock In \emph{The Thirty-eight Conference on Neural Information Processing Systems Datasets and Benchmarks Track}, 2024.

\bibitem[Chang et~al.(2024)Chang, Wang, Wang, Wu, Yang, Zhu, Chen, Yi, Wang, Wang, et~al.]{chang2024survey}
Yupeng Chang, Xu~Wang, Jindong Wang, Yuan Wu, Linyi Yang, Kaijie Zhu, Hao Chen, Xiaoyuan Yi, Cunxiang Wang, Yidong Wang, et~al.
\newblock A survey on evaluation of large language models.
\newblock \emph{ACM transactions on intelligent systems and technology}, 15\penalty0 (3):\penalty0 1--45, 2024.

\bibitem[Chen et~al.(2024)Chen, Liang, Sha, Shi, and Song]{chen2024hsr}
Bo~Chen, Yingyu Liang, Zhizhou Sha, Zhenmei Shi, and Zhao Song.
\newblock Hsr-enhanced sparse attention acceleration.
\newblock \emph{arXiv preprint arXiv:2410.10165}, 2024.

\bibitem[Cheng et~al.(2024)Cheng, Kolobov, Misra, Nie, and Swaminathan]{cheng2024llf}
Ching-An Cheng, Andrey Kolobov, Dipendra Misra, Allen Nie, and Adith Swaminathan.
\newblock Llf-bench: Benchmark for interactive learning from language feedback.
\newblock In \emph{ICLR 2024 Workshop on Large Language Model (LLM) Agents}, 2024.

\bibitem[Chhikara et~al.(2025)Chhikara, Khant, Aryan, Singh, and Yadav]{chhikara2025mem0}
Prateek Chhikara, Dev Khant, Saket Aryan, Taranjeet Singh, and Deshraj Yadav.
\newblock Mem0: Building production-ready ai agents with scalable long-term memory.
\newblock \emph{arXiv preprint arXiv:2504.19413}, 2025.

\bibitem[Child et~al.(2019)Child, Gray, Radford, and Sutskever]{child2019generating}
Rewon Child, Scott Gray, Alec Radford, and Ilya Sutskever.
\newblock Generating long sequences with sparse transformers.
\newblock \emph{arXiv e-prints}, pages arXiv--1904, 2019.

\bibitem[Dai et~al.(2019)Dai, Yang, Yang, Carbonell, Le, and Salakhutdinov]{dai2019transformer}
Zihang Dai, Zhilin Yang, Yiming Yang, Jaime Carbonell, Quoc~V Le, and Ruslan Salakhutdinov.
\newblock Transformer-xl: Attentive language models beyond a fixed-length context.
\newblock \emph{arXiv preprint arXiv:1901.02860}, 2019.

\bibitem[Dao et~al.(2022)Dao, Fu, Ermon, Rudra, and R{\'e}]{dao2022flashattention}
Tri Dao, Daniel~Y Fu, Stefano Ermon, Atri Rudra, and Christopher R{\'e}.
\newblock Flashattention: Fast and memory-efficient exact attention with io-awareness.
\newblock In \emph{Proceedings of the 35th Neural Information Processing Systems Conference (NeurIPS)}, 2022.

\bibitem[{DeepSeek-AI} et~al.(2025){DeepSeek-AI}, {Guo}, {Yang}, {Zhang}, {Song}, {Zhang}, {Xu}, {Zhu}, {Ma}, {Wang}, {Bi}, {Zhang}, {Yu}, {Wu}, {Wu}, {Gou}, {Shao}, {Li}, {Gao}, {Liu}, {Xue}, {Wang}, {Wu}, {Feng}, {Lu}, {Zhao}, {Deng}, {Zhang}, {Ruan}, {Dai}, {Chen}, {Ji}, {Li}, {Lin}, {Dai}, {Luo}, {Hao}, {Chen}, {Li}, {Zhang}, {Bao}, {Xu}, {Wang}, {Ding}, {Xin}, {Gao}, {Qu}, {Li}, {Guo}, {Li}, {Wang}, {Chen}, {Yuan}, {Qiu}, {Li}, {Cai}, {Ni}, {Liang}, {Chen}, {Dong}, {Hu}, {Gao}, {Guan}, {Huang}, {Yu}, {Wang}, {Zhang}, {Zhao}, {Wang}, {Zhang}, {Xu}, {Xia}, {Zhang}, {Zhang}, {Tang}, {Li}, {Wang}, {Li}, {Tian}, {Huang}, {Zhang}, {Wang}, {Chen}, {Du}, {Ge}, {Zhang}, {Pan}, {Wang}, {Chen}, {Jin}, {Chen}, {Lu}, {Zhou}, {Chen}, {Ye}, {Wang}, {Yu}, {Zhou}, {Pan}, {Li}, {Zhou}, {Wu}, {Ye}, {Yun}, {Pei}, {Sun}, {Wang}, {Zeng}, {Zhao}, {Liu}, {Liang}, {Gao}, {Yu}, {Zhang}, {Xiao}, {An}, {Liu}, {Wang}, {Chen}, {Nie}, {Cheng}, {Liu}, {Xie}, {Liu}, {Yang}, {Li}, {Su}, {Lin}, {Li}, {Jin}, {Shen}, {Chen}, {Sun}, {Wang},
  {Song}, {Zhou}, {Wang}, {Shan}, {Li}, {Wang}, {Wei}, {Zhang}, {Xu}, {Li}, {Zhao}, {Sun}, {Wang}, {Yu}, {Zhang}, {Shi}, {Xiong}, {He}, {Piao}, {Wang}, {Tan}, {Ma}, {Liu}, {Guo}, {Ou}, {Wang}, {Gong}, {Zou}, {He}, {Xiong}, {Luo}, {You}, {Liu}, {Zhou}, {Zhu}, {Xu}, {Huang}, {Li}, {Zheng}, {Zhu}, {Ma}, {Tang}, {Zha}, {Yan}, {Ren}, {Ren}, {Sha}, {Fu}, {Xu}, {Xie}, {Zhang}, {Hao}, {Ma}, {Yan}, {Wu}, {Gu}, {Zhu}, {Liu}, {Li}, {Xie}, {Song}, {Pan}, {Huang}, {Xu}, {Zhang}, and {Zhang}]{2025arXiv250112948D}
{DeepSeek-AI}, Daya {Guo}, Dejian {Yang}, Haowei {Zhang}, Junxiao {Song}, Ruoyu {Zhang}, Runxin {Xu}, Qihao {Zhu}, Shirong {Ma}, Peiyi {Wang}, Xiao {Bi}, Xiaokang {Zhang}, Xingkai {Yu}, Yu~{Wu}, Z.~F. {Wu}, Zhibin {Gou}, Zhihong {Shao}, Zhuoshu {Li}, Ziyi {Gao}, Aixin {Liu}, Bing {Xue}, Bingxuan {Wang}, Bochao {Wu}, Bei {Feng}, Chengda {Lu}, Chenggang {Zhao}, Chengqi {Deng}, Chenyu {Zhang}, Chong {Ruan}, Damai {Dai}, Deli {Chen}, Dongjie {Ji}, Erhang {Li}, Fangyun {Lin}, Fucong {Dai}, Fuli {Luo}, Guangbo {Hao}, Guanting {Chen}, Guowei {Li}, H.~{Zhang}, Han {Bao}, Hanwei {Xu}, Haocheng {Wang}, Honghui {Ding}, Huajian {Xin}, Huazuo {Gao}, Hui {Qu}, Hui {Li}, Jianzhong {Guo}, Jiashi {Li}, Jiawei {Wang}, Jingchang {Chen}, Jingyang {Yuan}, Junjie {Qiu}, Junlong {Li}, J.~L. {Cai}, Jiaqi {Ni}, Jian {Liang}, Jin {Chen}, Kai {Dong}, Kai {Hu}, Kaige {Gao}, Kang {Guan}, Kexin {Huang}, Kuai {Yu}, Lean {Wang}, Lecong {Zhang}, Liang {Zhao}, Litong {Wang}, Liyue {Zhang}, Lei {Xu}, Leyi {Xia}, Mingchuan {Zhang}, Minghua
  {Zhang}, Minghui {Tang}, Meng {Li}, Miaojun {Wang}, Mingming {Li}, Ning {Tian}, Panpan {Huang}, Peng {Zhang}, Qiancheng {Wang}, Qinyu {Chen}, Qiushi {Du}, Ruiqi {Ge}, Ruisong {Zhang}, Ruizhe {Pan}, Runji {Wang}, R.~J. {Chen}, R.~L. {Jin}, Ruyi {Chen}, Shanghao {Lu}, Shangyan {Zhou}, Shanhuang {Chen}, Shengfeng {Ye}, Shiyu {Wang}, Shuiping {Yu}, Shunfeng {Zhou}, Shuting {Pan}, S.~S. {Li}, Shuang {Zhou}, Shaoqing {Wu}, Shengfeng {Ye}, Tao {Yun}, Tian {Pei}, Tianyu {Sun}, T.~{Wang}, Wangding {Zeng}, Wanjia {Zhao}, Wen {Liu}, Wenfeng {Liang}, Wenjun {Gao}, Wenqin {Yu}, Wentao {Zhang}, W.~L. {Xiao}, Wei {An}, Xiaodong {Liu}, Xiaohan {Wang}, Xiaokang {Chen}, Xiaotao {Nie}, Xin {Cheng}, Xin {Liu}, Xin {Xie}, Xingchao {Liu}, Xinyu {Yang}, Xinyuan {Li}, Xuecheng {Su}, Xuheng {Lin}, X.~Q. {Li}, Xiangyue {Jin}, Xiaojin {Shen}, Xiaosha {Chen}, Xiaowen {Sun}, Xiaoxiang {Wang}, Xinnan {Song}, Xinyi {Zhou}, Xianzu {Wang}, Xinxia {Shan}, Y.~K. {Li}, Y.~Q. {Wang}, Y.~X. {Wei}, Yang {Zhang}, Yanhong {Xu}, Yao {Li}, Yao
  {Zhao}, Yaofeng {Sun}, Yaohui {Wang}, Yi~{Yu}, Yichao {Zhang}, Yifan {Shi}, Yiliang {Xiong}, Ying {He}, Yishi {Piao}, Yisong {Wang}, Yixuan {Tan}, Yiyang {Ma}, Yiyuan {Liu}, Yongqiang {Guo}, Yuan {Ou}, Yuduan {Wang}, Yue {Gong}, Yuheng {Zou}, Yujia {He}, Yunfan {Xiong}, Yuxiang {Luo}, Yuxiang {You}, Yuxuan {Liu}, Yuyang {Zhou}, Y.~X. {Zhu}, Yanhong {Xu}, Yanping {Huang}, Yaohui {Li}, Yi~{Zheng}, Yuchen {Zhu}, Yunxian {Ma}, Ying {Tang}, Yukun {Zha}, Yuting {Yan}, Z.~Z. {Ren}, Zehui {Ren}, Zhangli {Sha}, Zhe {Fu}, Zhean {Xu}, Zhenda {Xie}, Zhengyan {Zhang}, Zhewen {Hao}, Zhicheng {Ma}, Zhigang {Yan}, Zhiyu {Wu}, Zihui {Gu}, Zijia {Zhu}, Zijun {Liu}, Zilin {Li}, Ziwei {Xie}, Ziyang {Song}, Zizheng {Pan}, Zhen {Huang}, Zhipeng {Xu}, Zhongyu {Zhang}, and Zhen {Zhang}.
\newblock {DeepSeek-R1: Incentivizing Reasoning Capability in LLMs via Reinforcement Learning}.
\newblock \emph{arXiv e-prints}, art. arXiv:2501.12948, January 2025.
\newblock \doi{10.48550/arXiv.2501.12948}.

\bibitem[Ding et~al.(2023)Ding, Sheng, Chen, Mao, Peng, Chen, Liu, and Huo]{ding2023cervical}
Yu~Ding, Cong Sheng, Zhongyue Chen, Lingli Mao, Zhao Peng, Tingting Chen, Yiqun Liu, and Wanli Huo.
\newblock Cervical cancer segmentation based on full-scale feature fusion with cascading-attention and dilated convolution.
\newblock In \emph{Proc. of SPIE Vol}, volume 12800, pages 1280033--1, 2023.

\bibitem[Dong et~al.(2024)Dong, Tang, Li, Zhao, and Wen]{dong-etal-2024-bamboo}
Zican Dong, Tianyi Tang, Junyi Li, Wayne~Xin Zhao, and Ji-Rong Wen.
\newblock {BAMBOO}: A comprehensive benchmark for evaluating long text modeling capacities of large language models.
\newblock In Nicoletta Calzolari, Min-Yen Kan, Veronique Hoste, Alessandro Lenci, Sakriani Sakti, and Nianwen Xue, editors, \emph{Proceedings of the 2024 Joint International Conference on Computational Linguistics, Language Resources and Evaluation (LREC-COLING 2024)}, pages 2086--2099, Torino, Italia, May 2024. ELRA and ICCL.
\newblock URL \url{https://aclanthology.org/2024.lrec-main.188/}.

\bibitem[EducateMe(2024)]{educateme_knowledge_retention}
EducateMe.
\newblock Knowledge retention: 8 main strategies to improve it, February 2024.
\newblock URL \url{https://www.educate-me.co/blog/knowledge-retention-strategies}.

\bibitem[Gu and Dao(2024)]{gu2024mamba}
Albert Gu and Tri Dao.
\newblock Mamba: Linear-time sequence modeling with selective state spaces.
\newblock In \emph{First Conference on Language Modeling}, 2024.
\newblock URL \url{https://openreview.net/forum?id=tEYskw1VY2}.

\bibitem[{Guo} et~al.(2025){Guo}, {Liu}, {Heng}, {Tse-Hsun}, {Chen}, and {Wang}]{2025arXiv250523596G}
Linqiang {Guo}, Wei {Liu}, Yi~Wen {Heng}, {Tse-Hsun}, {Chen}, and Yang {Wang}.
\newblock {MAPLE: A Mobile Agent with Persistent Finite State Machines for Structured Task Reasoning}.
\newblock \emph{arXiv e-prints}, art. arXiv:2505.23596, May 2025.
\newblock \doi{10.48550/arXiv.2505.23596}.

\bibitem[Guti\'{e}rrez et~al.(2024)Guti\'{e}rrez, Shu, Gu, Yasunaga, and Su]{NEURIPS2024_6ddc001d}
Bernal~Jim\'{e}nez Guti\'{e}rrez, Yiheng Shu, Yu~Gu, Michihiro Yasunaga, and Yu~Su.
\newblock Hipporag: Neurobiologically inspired long-term memory for large language models.
\newblock In A.~Globerson, L.~Mackey, D.~Belgrave, A.~Fan, U.~Paquet, J.~Tomczak, and C.~Zhang, editors, \emph{Advances in Neural Information Processing Systems}, volume~37, pages 59532--59569. Curran Associates, Inc., 2024.
\newblock URL \url{https://proceedings.neurips.cc/paper_files/paper/2024/file/6ddc001d07ca4f319af96a3024f6dbd1-Paper-Conference.pdf}.

\bibitem[Han et~al.(2024)Han, Chen, Du, Kong, Xiong, and Pan]{han2024layer}
Yi~Han, Hang Chen, Jun Du, Chang-Qing Kong, Shi-fu Xiong, and Jia Pan.
\newblock Layer-adaptive low-rank adaptation of large asr model for low-resource multilingual scenarios.
\newblock In \emph{2024 IEEE 14th International Symposium on Chinese Spoken Language Processing (ISCSLP)}, pages 696--700. IEEE, 2024.

\bibitem[{Hao} et~al.(2024){Hao}, {Han}, {Jiang}, {Li}, {Wu}, {Zhong}, {Zhou}, and {Tang}]{2024arXiv240101629H}
Shuang {Hao}, Wenfeng {Han}, Tao {Jiang}, Yiping {Li}, Haonan {Wu}, Chunlin {Zhong}, Zhangjun {Zhou}, and He~{Tang}.
\newblock {Synthetic Data in AI: Challenges, Applications, and Ethical Implications}.
\newblock \emph{arXiv e-prints}, art. arXiv:2401.01629, January 2024.
\newblock \doi{10.48550/arXiv.2401.01629}.

\bibitem[Hsieh et~al.(2024)Hsieh, Sun, Kriman, Acharya, Rekesh, Jia, Zhang, and Ginsburg]{hsieh2024ruler}
Cheng-Ping Hsieh, Simeng Sun, Samuel Kriman, Shantanu Acharya, Dima Rekesh, Fei Jia, Yang Zhang, and Boris Ginsburg.
\newblock Ruler: What's the real context size of your long-context language models?
\newblock \emph{CoRR}, 2024.

\bibitem[Kamradt(2023)]{needle-in-haystack}
Gregory Kamradt.
\newblock Needle in a haystack - pressure testing llms, 2023.
\newblock URL \url{https://github.com/gkamradt/LLMTest_NeedleInAHaystack}.

\bibitem[Kitaev et~al.(2020)Kitaev, Kaiser, and Levskaya]{kitaevreformer}
Nikita Kitaev, Lukasz Kaiser, and Anselm Levskaya.
\newblock Reformer: The efficient transformer.
\newblock In \emph{International Conference on Learning Representations}, 2020.

\bibitem[Kuratov et~al.(2024)Kuratov, Bulatov, Anokhin, Rodkin, Sorokin, Sorokin, and Burtsev]{kuratov2024babilong}
Yury Kuratov, Aydar Bulatov, Petr Anokhin, Ivan Rodkin, Dmitry Sorokin, Artyom Sorokin, and Mikhail Burtsev.
\newblock Babilong: Testing the limits of llms with long context reasoning-in-a-haystack.
\newblock \emph{Advances in Neural Information Processing Systems}, 37:\penalty0 106519--106554, 2024.

\bibitem[Kwon et~al.(2023)Kwon, Li, Zhuang, Sheng, Zheng, Yu, Gonzalez, Zhang, and Stoica]{kwon2023efficient}
Woosuk Kwon, Zhuohan Li, Siyuan Zhuang, Ying Sheng, Lianmin Zheng, Cody~Hao Yu, Joseph Gonzalez, Hao Zhang, and Ion Stoica.
\newblock Efficient memory management for large language model serving with pagedattention.
\newblock In \emph{Proceedings of the 29th Symposium on Operating Systems Principles}, pages 611--626, 2023.

\bibitem[{Lee} et~al.(2024){Lee}, {Chen}, {Furuta}, {Canny}, and {Fischer}]{2024arXiv240209727L}
Kuang-Huei {Lee}, Xinyun {Chen}, Hiroki {Furuta}, John {Canny}, and Ian {Fischer}.
\newblock {A Human-Inspired Reading Agent with Gist Memory of Very Long Contexts}.
\newblock \emph{arXiv e-prints}, art. arXiv:2402.09727, feb 2024.
\newblock \doi{10.48550/arXiv.2402.09727}.

\bibitem[Li et~al.(2023)Li, Wang, Zheng, and Zhang]{li2023loogle}
Jiaqi Li, Mengmeng Wang, Zilong Zheng, and Muhan Zhang.
\newblock Loogle: Can long-context language models understand long contexts?
\newblock \emph{arXiv e-prints}, pages arXiv--2311, 2023.

\bibitem[Liu et~al.(2023)Liu, Zaharia, and Abbeel]{liuringattention}
Hao Liu, Matei Zaharia, and Pieter Abbeel.
\newblock Ringattention with blockwise transformers for near-infinite context.
\newblock In \emph{The Twelfth International Conference on Learning Representations}, 2023.

\bibitem[{Liu} et~al.(2024){Liu}, {Wei}, {Liu}, {Si}, {Zhang}, {Rao}, {Zheng}, {Peng}, {Yang}, {Zhou}, and {Dai}]{2024arXiv240407503L}
Ruibo {Liu}, Jerry {Wei}, Fangyu {Liu}, Chenglei {Si}, Yanzhe {Zhang}, Jinmeng {Rao}, Steven {Zheng}, Daiyi {Peng}, Diyi {Yang}, Denny {Zhou}, and Andrew~M. {Dai}.
\newblock {Best Practices and Lessons Learned on Synthetic Data}.
\newblock \emph{arXiv e-prints}, art. arXiv:2404.07503, April 2024.
\newblock \doi{10.48550/arXiv.2404.07503}.

\bibitem[Liu et~al.(2024)Liu, Yu, Zhang, Xu, Lei, Lai, Gu, Ding, Men, Yang, et~al.]{liu2024agentbench}
Xiao Liu, Hao Yu, Hanchen Zhang, Yifan Xu, Xuanyu Lei, Hanyu Lai, Yu~Gu, Hangliang Ding, Kaiwen Men, Kejuan Yang, et~al.
\newblock Agentbench: Evaluating llms as agents.
\newblock In \emph{ICLR}, 2024.

\bibitem[OpenAI(2024)]{openai_gpt_4o}
OpenAI.
\newblock Hello gpt-4o, 2024.
\newblock URL \url{https://openai.com/index/hello-gpt-4o/}.

\bibitem[Packer et~al.(2023)Packer, Fang, Patil, Lin, Wooders, and Gonzalez]{packer2023memgpt}
Charles Packer, Vivian Fang, Shishir\_G Patil, Kevin Lin, Sarah Wooders, and Joseph\_E Gonzalez.
\newblock Memgpt: Towards llms as operating systems.
\newblock \emph{arXiv}, 2023.

\bibitem[Pagliardini et~al.(2023)Pagliardini, Paliotta, Jaggi, and Fleuret]{pagliardini2023faster}
Matteo Pagliardini, Daniele Paliotta, Martin Jaggi, and Fran{\c{c}}ois Fleuret.
\newblock Faster causal attention over large sequences through sparse flash attention.
\newblock \emph{arXiv e-prints}, pages arXiv--2306, 2023.

\bibitem[Peng et~al.(2023)Peng, Alcaide, Anthony, Albalak, Arcadinho, Biderman, Cao, Cheng, Chung, Derczynski, et~al.]{peng2023rwkv}
Bo~Peng, Eric Alcaide, Quentin~Gregory Anthony, Alon Albalak, Samuel Arcadinho, Stella Biderman, Huanqi Cao, Xin Cheng, Michael~Nguyen Chung, Leon Derczynski, et~al.
\newblock Rwkv: Reinventing rnns for the transformer era.
\newblock In \emph{The 2023 Conference on Empirical Methods in Natural Language Processing}, 2023.

\bibitem[Rae and Razavi(2020)]{rae2020transformers}
Jack Rae and Ali Razavi.
\newblock Do transformers need deep long-range memory?
\newblock In \emph{Proceedings of the 58th Annual Meeting of the Association for Computational Linguistics}, pages 7524--7529. Association for Computational Linguistics, 2020.

\bibitem[Shaham et~al.(2023)Shaham, Ivgi, Efrat, Berant, and Levy]{shaham2023zeroscrolls}
Uri Shaham, Maor Ivgi, Avia Efrat, Jonathan Berant, and Omer Levy.
\newblock Zeroscrolls: A zero-shot benchmark for long text understanding.
\newblock In \emph{2023 Findings of the Association for Computational Linguistics: EMNLP 2023}, pages 7977--7989. Association for Computational Linguistics (ACL), 2023.

\bibitem[{Shan} et~al.(2025){Shan}, {Luo}, {Zhu}, {Yuan}, and {Wu}]{2025arXiv250402441S}
Lianlei {Shan}, Shixian {Luo}, Zezhou {Zhu}, Yu~{Yuan}, and Yong {Wu}.
\newblock {Cognitive Memory in Large Language Models}.
\newblock \emph{arXiv e-prints}, art. arXiv:2504.02441, April 2025.
\newblock \doi{10.48550/arXiv.2504.02441}.

\bibitem[Sun et~al.(2022)Sun, Thai, and Iyyer]{sun2022chapterbreak}
Simeng Sun, Katherine Thai, and Mohit Iyyer.
\newblock Chapterbreak: A challenge dataset for long-range language models.
\newblock In \emph{Proceedings of the 2022 Conference of the North American Chapter of the Association for Computational Linguistics: Human Language Technologies}, pages 3704--3714, 2022.

\bibitem[{Wu} et~al.(2025){Wu}, {Liang}, {Zhang}, {Wang}, {Zhang}, {Guo}, {Tang}, and {Liu}]{2025arXiv250415965W}
Yaxiong {Wu}, Sheng {Liang}, Chen {Zhang}, Yichao {Wang}, Yongyue {Zhang}, Huifeng {Guo}, Ruiming {Tang}, and Yong {Liu}.
\newblock {From Human Memory to AI Memory: A Survey on Memory Mechanisms in the Era of LLMs}.
\newblock \emph{arXiv e-prints}, art. arXiv:2504.15965, April 2025.
\newblock \doi{10.48550/arXiv.2504.15965}.

\bibitem[{Xie} et~al.(2025){Xie}, {Harel}, {Ran}, {Li}, {Li}, {Yang}, {Wang}, {Chen}, {Zhang}, {Zhang}, {Li}, {Zhang}, {Li}, and {Marron}]{2025arXiv250107487X}
Tao {Xie}, David {Harel}, Dezhi {Ran}, Zhenwen {Li}, Maoliang {Li}, Zhi {Yang}, Leye {Wang}, Xiang {Chen}, Ying {Zhang}, Wentao {Zhang}, Meng {Li}, Chen {Zhang}, Linyi {Li}, and Assaf {Marron}.
\newblock {Data and System Perspectives of Sustainable Artificial Intelligence}.
\newblock \emph{arXiv e-prints}, art. arXiv:2501.07487, January 2025.
\newblock \doi{10.48550/arXiv.2501.07487}.

\bibitem[Xu et~al.(2022)Xu, Gou, Wu, Niu, Wu, Wang, and Wang]{xu2022long}
Xinchao Xu, Zhibin Gou, Wenquan Wu, Zheng-Yu Niu, Hua Wu, Haifeng Wang, and Shihang Wang.
\newblock Long time no see! open-domain conversation with long-term persona memory.
\newblock \emph{Findings of the Association for Computational Linguistics: ACL 2022}, 2022.

\bibitem[Yu et~al.(2024)Yu, Qin, Fu, Huang, Xu, Liu, and Liu]{MemoryScope}
Li~Yu, Tiancheng Qin, Qingxu Fu, Sen Huang, Xianzhe Xu, Zhaoyang Liu, and Boyin Liu.
\newblock Memoryscope, 09 2024.
\newblock URL \url{https://github.com/modelscope/MemoryScope}.
\newblock GitHub repository.

\bibitem[Zhang et~al.(2025)Zhang, Dai, Wu, Yang, Wang, Tang, and Liu]{zhang2025survey}
Chen Zhang, Xinyi Dai, Yaxiong Wu, Qu~Yang, Yasheng Wang, Ruiming Tang, and Yong Liu.
\newblock A survey on multi-turn interaction capabilities of large language models.
\newblock \emph{arXiv e-prints}, pages arXiv--2501, 2025.

\bibitem[Zhang et~al.(2024)Zhang, Chen, Hu, Xu, Chen, Hao, Han, Thai, Wang, Liu, et~al.]{zhang2024bench}
Xinrong Zhang, Yingfa Chen, Shengding Hu, Zihang Xu, Junhao Chen, Moo~Khai Hao, Xu~Han, Zhen~Leng Thai, Shuo Wang, Zhiyuan Liu, et~al.
\newblock \newblock bench: Extending long context evaluation beyond 100k tokens.
\newblock \emph{CoRR}, 2024.

\bibitem[Zhong et~al.(2023)Zhong, Guo, Gao, Ye, and Wang]{zhong2023memorybank}
Wanjun Zhong, Lianghong Guo, Qiqi Gao, He~Ye, and Yanlin Wang.
\newblock Memorybank: Enhancing large language models with long-term memory.
\newblock \emph{CoRR}, 2023.

\bibitem[Zhou et~al.(2023)Zhou, Xu, Zhu, Zhou, Lo, Sridhar, Cheng, Ou, Bisk, Fried, et~al.]{zhouwebarena}
Shuyan Zhou, Frank~F Xu, Hao Zhu, Xuhui Zhou, Robert Lo, Abishek Sridhar, Xianyi Cheng, Tianyue Ou, Yonatan Bisk, Daniel Fried, et~al.
\newblock Webarena: A realistic web environment for building autonomous agents.
\newblock In \emph{The Twelfth International Conference on Learning Representations}, 2023.

\end{thebibliography}
\normalsize  

\medskip

\end{document}